\documentclass[conference]{IEEEtran}
\IEEEoverridecommandlockouts
\usepackage{cite}
\usepackage{amsmath,amssymb,amsfonts}
\usepackage{algorithmic}
\usepackage{graphicx}
\usepackage[normalem]{ulem}
\useunder{\uline}{\ul}{}
\usepackage{textcomp}
\usepackage{xcolor}
\usepackage{caption}
\usepackage{subcaption}
\usepackage{multirow}
\usepackage{booktabs}
\def\BibTeX{{\rm B\kern-.05em{\sc i\kern-.025em b}\kern-.08em
    T\kern-.1667em\lower.7ex\hbox{E}\kern-.125emX}}
\begin{document}

\title{Multi-Task Time Series Forecasting With Shared Attention}

\author{\IEEEauthorblockN{Zekai Chen\IEEEauthorrefmark{1},
Jiaze E\IEEEauthorrefmark{1},
Xiao Zhang\IEEEauthorrefmark{2}, 
Hao Sheng\IEEEauthorrefmark{3} and
Xiuzheng Cheng\IEEEauthorrefmark{1}}
\IEEEauthorblockA{\IEEEauthorrefmark{1}Department of Computer Science\\
George Washington University,
Washington, DC, USA\\ \{zech\_chan, ejiaze, cheng\}@gwu.edu}
\IEEEauthorblockA{\IEEEauthorrefmark{2}School of Computer Science and Technology, Shandong University, China\\
xiaozhang@sdu.edu.cn}
\IEEEauthorblockA{\IEEEauthorrefmark{3}
School of Computing Science and Engineering, Beihang University, China\\shenghao@buua.edu.cn}}


\maketitle

\begin{abstract}
Time series forecasting is a key component in many industrial and business decision processes and recurrent neural network (RNN) based models have achieved impressive progress on various time series forecasting tasks.  However, most of the existing methods focus on single-task forecasting problems by learning separately based on limited supervised objectives, which often suffer from insufficient training instances. As the Transformer architecture and other attention-based models have demonstrated its great capability of capturing long term dependency, we propose two self-attention based sharing schemes for multi-task time series forecasting which can train jointly across multiple tasks. We augment a sequence of paralleled Transformer encoders with an external public multi-head attention function, which is updated by all data of all tasks. Experiments on a number of real-world multi-task time series forecasting tasks show that our proposed architectures can not only outperform the state-of-the-art single-task forecasting baselines but also outperform the RNN-based multi-task forecasting method. 
\end{abstract}


\section{Introduction}
\begin{figure}
    \centering
    \includegraphics[width=.75\linewidth]{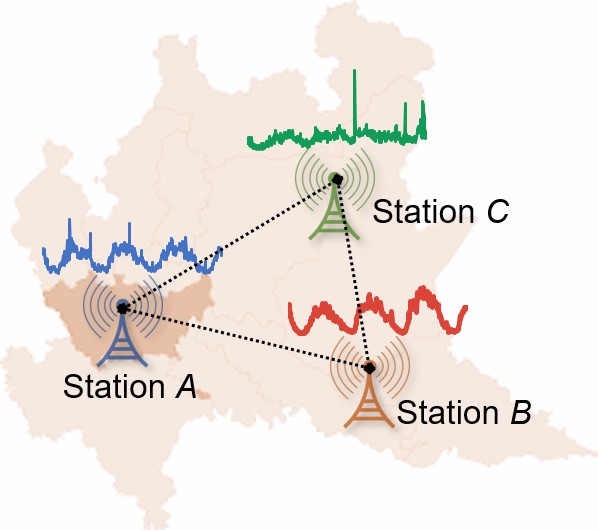}
    \caption{A paradigm of cellular traffic data collected from multiple base stations.}
    \label{fig:station}
\end{figure}

Multi-task time series forecasting, i.e. the prediction of multiple time series data from different tasks, is a crucial problem within both time series forecasting and multi-task learning.  In contrast to single-task learning, multi-task time series forecasts provide users with access to estimates across multiple related time series paths, allowing them to optimize their actions in multiple related domains simultaneously in the future. The development of multi-task time series forecasting can benefit many applications such as stock prices forecasting, weather forecasting, business planning, traffic prediction, resources allocation, optimization in IoT and many others. Especially in recent years, with the rapid development of the Internet of Things (IoT), billions of connected mobile devices have generated massive data and further bring many novel applications that can change human life \cite{Morales2016, Sun2016}. Analyzing these data appropriately can bring considerable socio-economic benefits such as target-advertising based on accurate prediction of cellular traffic data, real-time health status monitoring, etc. Different from general single-task forecasting problems, practical multi-task forecasting applications commonly have access to a variety of data collection resources as shown in Fig. \ref{fig:station}. In this cellular traffic forecasting problem, all the base stations are well deployed in certain urban areas. Station $A$ and Station $B$ share a similar pattern possibly due to geographical proximity while different from the traffic pattern of Station $C$ a lot. If we want to forecast the future cellular traffic of any of them, one main challenge is that how we can fully utilize both commonality and difference among these time series from different stations with the aim of mutual benefit. It is vital especially when there is little acquired data from each station due to failure or privacy reasons.   
Traditional time series forecasting methods include auto-regressive integrated moving average (ARIMA) \cite{Asteriou2016, Xu2017}, vector auto-regression (VAR) \cite{Evgeniou2004}, support vector regression (SVR) \cite{Drucker1997}, etc. Recently, deep neural networks \cite{Graves2013, Sutskever2014, Oord2016, Luong2016} offers an alternative. The recurrent neural networks (RNNs) have become one of the most popular models in sequence modeling research. Two variants of RNN in particular, the long short term memory (LSTM) \cite{Hochreiter1997} and the gated recurrent unit (GRU) \cite{Chung2014}, have significantly improved the state-of-the-art performance in time series forecasting and other sequence modeling tasks. Especially, meta multi-task learning \cite{Liu2016a, Liu2016, Chen2018} proposed a new sharing scheme of composition function across multiple tasks based on LSTM models. Most recently, as the ability to capture long term dependency with good parallelism, the Transformer architecture \cite{Vaswani2017, Li2019} has been widely used in natural language processing (NLP) and yields state-of-the-art results on a number of tasks. Despite the popularity of various sequence modeling research, most of the work focus on either single-task learning or combining multi-task learning with recurrent neural networks and there have been few works in combining MTL with Transformer, especially the self-attention mechanism. 

In this paper, we propose to bridge the gap between multi-task learning and Transformer attention-based architectures by designing a shared-private attention sharing scheme MTL-Trans to jointly train on multiple related tasks. Inspired by shared external memory \cite{Liu2016} based on LSTM models, we propose two architectures of sharing attention information among different tasks under a multi-task learning framework. All the related tasks are integrated into a single system that is trained jointly. Specifically, we use an external multi-head attention function as a shared attention layer to store long-term self-attention information and knowledge across different related tasks. 

We demonstrate the effectiveness of our architectures on a real-world multi-task time series forecasting task. Experimental results show that jointly learning of multiple related tasks can improve the performance of each task relative to learning them independently. Additionally, attention-based sharing architectures can outperform the RNN-based sharing architectures. In summary:
\begin{itemize}
    \item We are the first to propose an attention-based multi-task learning framework (MTL-Trans) to solve multi-task time series forecasting problems.
    \item We propose two different attention sharing architectures for sharing self-attention information among different tasks during jointly training process. The external public multi-head attention helps to capture and recording self-attention information across different tasks.
    \item We conducted extensive experiments on a real-world multi-task time series forecasting task, and the proposed approach obtains signiﬁcant improvement over state-of-the-art baseline methods.
\end{itemize}

\section{Related Work}
\begin{figure*}[htbp]
    \centering
    \includegraphics[width=.8\linewidth]{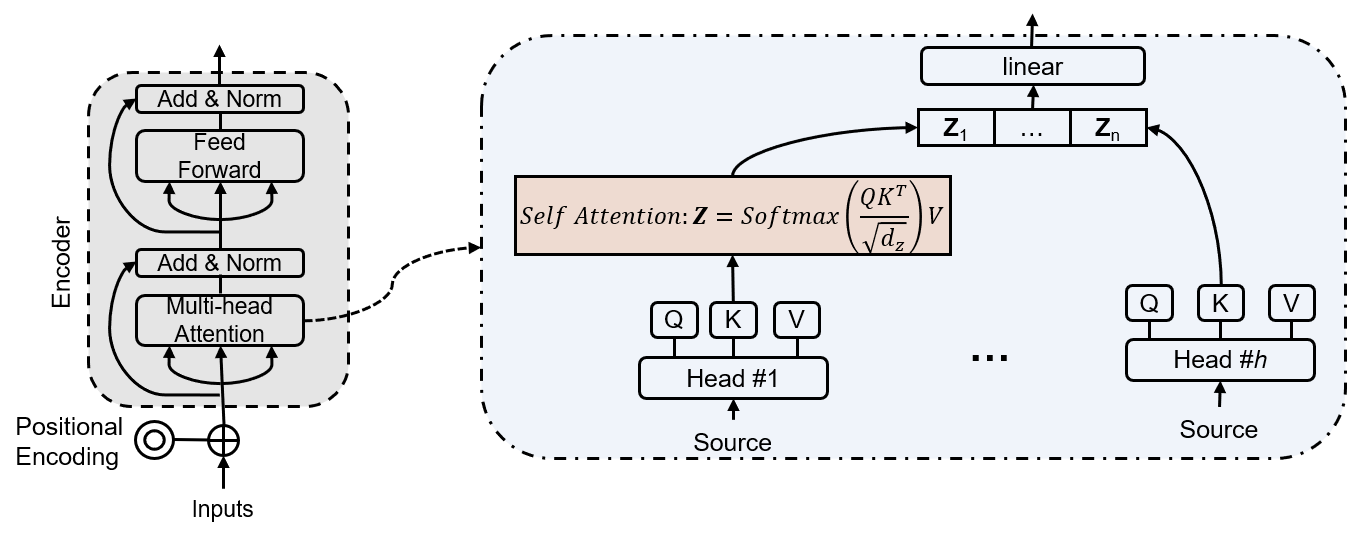}
    \caption{Multi-head attention architecture in Transformer.}
    \label{fig:classic_attention}
\end{figure*}

\textbf{Time Series Forecasting.} Even though forecasting can be considered as a subset of supervised regression problems, some specific tools are necessary due to the temporal nature of observations. Traditional data-driven approaches such as auto-regressive integrated moving average (ARIMA) \cite{Asteriou2016, Xu2017} model, Kalman filtering \cite{Xu2017}, support vector regression (SVR) \cite{Drucker1997}, and holt-winters exponential smoothing \cite{Tikunov2007} remain popular. Also, with the rise of various deep learning techniques, many efficient deep models have been proposed for time series forecasting. The recurrent neural networks (RNNs) \cite{Hochreiter1997, Graves2013, Sutskever2014, Lai2018} are powerful tools to model the temporal sequence data. Specifically, based on the variational auto-encoder (VAE) framework \cite{Ya2007, Kingma2014}, several variants of the RNNs have been proposed to process a highly structured natural sequence by capturing long-term dependencies. DCRNN \cite{Li2018} proposed a deep learning framework for traffic forecasting that incorporates both spatial and temporal dependency in the time serial traffic flow. DSSM \cite{Rangapuram2018} presented a probabilistic way that combined state-space models with a recurrent neural network. DeepAR \cite{Salinas2019} estimated a time series' future probability distribution given its past by training an auto-regressive recurrent neural network model. 

\textbf{Transformer framework.} Even though the problems of gradient vanishing or explosion have been overcome by LSTMs to some extent, the RNN based models are still not able to modeling very long term dependency \cite{Hochreiter1997}. Self-attention, also known as intra-attention, is an attention mechanism relating different positions of a single sequence in order to compute a representation of the same sequence. It has been shown to be very useful in machine reading \cite{Cheng2016}, abstractive summarization, or image description generation. With the help of the attention mechanism \cite{Bahdanau2015, Luong2015, Vaswani2017}, the dependencies between source and target sequences are not restricted by the in-between distance anymore. Among all the attention based variants, the Transformer model \cite{Vaswani2017} emerges as one of the most effective paradigms for dealing with long-term sequence modeling. It presented a lot of improvements to the soft attention \cite{Xu2015} and make it possible to do sequence to sequence modeling without recurrent network units. The proposed “transformer” model is entirely built on the self-attention mechanisms without using sequence-aligned recurrent architecture. Recently, temporal fusion transformer \cite{Lim2019} combines high-performance multi-horizon forecasting with interpretable insights into temporal dynamics, which further demonstrated the advantages of attention mechanism in time sequence forecasting. However, most existing research approaches focus on the single-task learning problem. When faced with multiple time series sequences collected from many other related domains, the existing models have to train each task separately without a strong multi-task generalization capability. 

\textbf{Multi-task Learning.} Multi-task learning (MTL) is an important machine learning paradigm that aims at improving the generalization performance of a task using other related tasks \cite{Evgeniou2004, Ya2007, Kim2010, Graves2013}. Particularly, CellScope \cite{Iyer2018} applied multi-task learning to resolve the trade-off between data collection latency and analysis accuracy in real-time mobile data analytic, in which data from geographically nearby base stations were grouped together. Luong et al. \cite{Luong2016} examined three multi-task strategies for sequence to sequence models: the $one$-$to$-$many$ setting, the $many$-$to$-$one$ setting and the $many$-$to$-$many$ setting. Liu et al. \cite{Liu2016a, Liu2016, Chen2018} proposed several multi-task sequence learning architectures by using enhanced and external memory to share information among paralleled RNN models. Despite the wide interest of various sequence modeling research, there is hardly any previous work done on combining multi-task time series forecasting with attention based architectures based on my knowledge. 

\section{Shared-Private Attention Sharing Scheme}
\label{section:scheme}

\subsection{Task Definition}
In this work, we focus on single-step forecasting. The basic definition of a multi-task time series single-step forecasting problem is: Given a dataset $\mathcal{D} = \{ \{ {\mathbf{x}_{mn}},{\mathbf{y}_{mn}}\} |_{n = 1}^{{N_m}}\} |_{m = 1}^M$ with multiple sequence tasks, $M$ denotes the number of tasks, $N_m$ means the number of instances in $m$-th task, $\mathbf{x}_{mn}$ is the $n$-th sample in $m$-th task. 
For instance, ${\mathbf{x}_{mn}} = \{ x_{mn}^{t_1}, \ldots ,x_{mn}^{t_s}\}$ could be historical observation values with length $s$, and ${\mathbf{y}_{mn}} = \{ y_{mn}^{t_{2}}, \ldots$ $ ,y_{mn}^{t_{s+1}}\}$ means the future time series sequence with the same length $s$ corresponding to ${\mathbf{x}_{mn}}$. 
Hence, the goal of the multiple single-step time series forecasting tasks is to learn a function that maps observation sequence $\{ {\mathbf{x}_{mn}}|_{n = 1}^{{N_m}}\} |_{m = 1}^M$ to future sequence $\{ {\mathbf{y}_{mn}}|_{n = 1}^{{N_m}}\} |_{m = 1}^M$: $f({\mathbf{x}_{mn}}) \to {\mathbf{y}_{mn}}$ jointly by utilizing the latent similarities among the tasks based on multi-task learning. 

\subsection{Preliminary Exploration}
\begin{figure*}[htbp]
\centerline{\includegraphics[width=.75\linewidth]{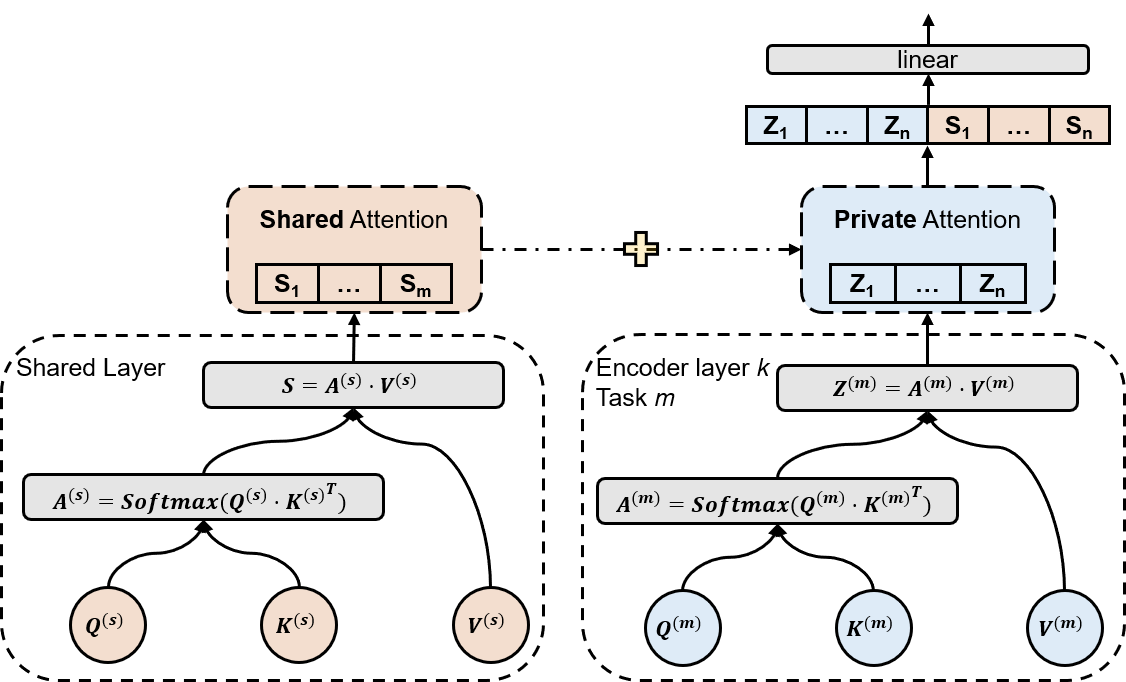}}
\caption{A global Shared-Private multi-head attention scheme for multi-task learning.}
\label{fig:share_attention}
\end{figure*}
\textbf{Scaled Dot-Product Attention.} The original Transformer used a particular scaled dot-product attention \cite{Vaswani2017}. The input consists of queries and keys of dimension $d_k$, and values of dimension $d_v$. The dot product of the query with all keys would be computed and divided each by $\sqrt{d_k}$. A softmax function would be applied to obtain the weights on the values. In practice, the attention function on a set of queries is computed simultaneously by being packed together into a matrix $Q$. The keys and values are also packed together into matrices $K$ and $V$, as a result, the matrix of outputs is as following:
\begin{equation}
    \text{Attention}(Q, K, V) = \text{softmax}(\frac{QK^T}{\sqrt{d_k}})V
\end{equation}

More specifically, this attention mechanism operates on an input sequence, $\mathbf{x} = (x_1, x_2, \cdots, x_n)$ with $n$ elements where $x_i\in \mathcal{R}^{d_x}$, and computes a new sequence $\mathbf{z}=(z_1, z_2, \cdots, z_n)$ of the same length where $z_i\in \mathcal{R}^{d_z}$. 

Each output element, $z_i$, is computed as weighted sum of a linearly transformed input elements:
\begin{equation}
z_{i}=\sum_{j=1}^{n} \alpha_{i j}\left(x_{j} W^{V}\right)
\end{equation}

Each weight coefficient, $\alpha_{ij}$, is computed using a softmax funtion:
\begin{equation}
\alpha_{i j}=\frac{\exp e_{i j}}{\sum_{k=1}^{n} \exp e_{i k}}
\end{equation}

And $e_{ij}$ is computed by the attention function that essentially finds the similarity between queries and keys using this dot-product so as to perform a soft-addressing process:
\begin{equation}e_{i j}=\frac{\left(x_{i} W^{Q}\right)\left(x_{j} W^{K}\right)^{T}}{\sqrt{d_{z}}}\end{equation}
where $W^Q\in \mathcal{R}^{d_x\times d_k}$, $W^K\in \mathcal{R}^{d_x\times d_k}$, $W^V\in \mathcal{R}^{d_x\times d_v}$ are parameter matrices. In practice, we usually set $d_k=d_v=d_z$.

\textbf{Multi-head Attention.} Instead of performing a single attention function with $d_{model}$-dimension keys, values, and queries, it is beneficial to linearly project the queries, keys, and values $h$ times with different, learned linear projections to $d_k$, $d_k$ and $d_v$ dimensions, respectively. Parallel attention function can be performed on each of these projected versions of queries, keys, and values, yielding $d_v$-dimensional output values. These are concatenated and once again projected, resulting in the final values. This multi-head attention mechanism (MHA) allows the model to jointly attend to information from different representation subspaces at different positions. 

Generally, once we capture the new sequences output from the multi-head functions as $\mathbf{z}^{(1)}, \mathbf{z}^{(2)}, \cdots, \mathbf{z}^{(h)}$ where $z^{(i)}$ means the attention score computed by the $i$th head. We concatenate these scores as $[\mathbf{z}^{(1)}\mathbf{z}^{(2)}\cdots \mathbf{z}^{(h)}]$ and multiple them with an additional weight matrix to align the dimension with targets. See Fig. \ref{fig:classic_attention} for an illustration of the multi-head attention model used in Transformer.

\textbf{Masking Self-Attention Heads.} In order to prevent from attending to subsequent positions, we apply attention masks, combined with the fact that the output embeddings are offset by one position,  ensuring that the predictions for position $i$ can depend only on the known outputs at positions before $i$.

\textbf{Shared-Private Attention Scheme.} The main challenge of multi-task learning is how to design the sharing scheme. Despite the big success of recurrent neural networks in temporal pattern recognition, long-term information has to sequentially travel through all cells before getting to the present processing cell which means it can be easily corrupted by being multiplied much time by small negative numbers. This is the major cause of shared information forgetting. Fortunately, the Transformer helps drawing global dependencies between inputs and outputs by creatively relies entirely on the attention mechanism result in setting the distance between any two elements in a sequence to 1. Additionally, its good parallelism is well suited for multi-task learning. In this paper, we plan to provide a shared attention model MTL-Trans among multiple tasks based on the Transformer with two different sharing architectures.

\subsection{General Global Shared Attention}
Though the classic Transformer model employs an encoder-decoder structure, consisting of stacked encoder and decoder layers, in this work, we only consider the self-attention without giving concern to the encoder-decoder attention since our work focuses on a sequence self-modeling process. To exploit the shared information between different tasks, the general global shared attention architecture consists of private (task-specific) encoder layers and a shared (task-invariant) attention layer. The shared multi-head attention layer captures the shared information for all the tasks. In this architecture, the source time series is modeled by task-specific stacked self-attention based encoders. More formally, given an input time series sequence $\mathbf{x}^{(m)} = (x_1, x_2, \cdots, x_n)$ from a random selected task $m$, the shared attention information output $\mathbf{s}^{(m)} = (s_1, s_2, \cdots, s_n)$ from the public multi-head attention layer is defined as
\begin{equation}
    \mathbf{s}^{(m)} = \text{MultiheadAttention}_{shared}(\mathbf{x}^{(m)})
\end{equation}
where $s_i\in \mathcal{R}^{d_s}$. Simultaneously, the task-specific attention output $\mathbf{z}^{(m)}_{k}=(z_1, z_2, \cdots, z_n)$ of multi-head attention from the $k$th encoder layer is computed as
\begin{equation}
    \mathbf{z}^{(m)}_{k} = \text{MultiheadAttention}_{k}(\mathbf{z}^{(m)}_{k-1})
\end{equation} 
where $\mathbf{z}^{(m)}_{k-1}$ is the output of the $(k-1)$th encoder from task $m$.
The shared attention values and private values are then arranged in concatenated manner. The task-specific encoders take the output of the shared layer as input. The attention output from $k$th encoder layer is updated as
\begin{equation}
    \mathbf{z}^{(m)}_{k} = 
    \begin{bmatrix}
    \mathbf{z}^{(m)}_{k}\\ 
    \mathbf{s}^{(m)}
    \end{bmatrix}^{T}W^{O}
\end{equation}
where $W^{O}\in \mathcal{R}^{(d_s + d_z)\times d_z}$ is a parameter matrix that computes the weighted average information on a combination of both shared attention and private attention. This also helps align the outputs as the same dimension with our target sequences. The output is then fed into a fully connected feed-forward network (FFN) just as the original Transformer does. See Fig. \ref{fig:share_attention} for the illustration of a general global attention sharing scheme.

\subsection{Hybrid Local-global Shared Attention}

Different from the general global attention sharing scheme, a hybrid local-global shared attention mechanism can make all tasks share a global attention memory, but can also record task-specific information besides shared information. 

More generally, given an output sequence $\mathbf{z}^{(m)}_{k}=(z_1, z_2, \cdots, z_n)$ from the $k$th encoder layer for a random task $m$. The output will be fed back into the shared multi-head attention layer defined as
\begin{equation}
    \mathbf{s}^{(m)}_{updated} = \text{MultiheadAttention}_{shared}(\mathbf{z}^{(m)}_{k})
\end{equation}

Again, the shared attention values and private outputs are arranged in concatenated manner and fed into the next encoder layer. The multi-head attention output from $(k+1)$th encoder layer is finally as
\begin{equation}
    \mathbf{z}^{(m)}_{k+1} = \text{MultiheadAttention}_{k+1}(\begin{bmatrix}
    \mathbf{z}^{(m)}_{k}\\
    \mathbf{s}^{(m)}_{updated}
    \end{bmatrix})
\end{equation}

\begin{figure}[htbp]
    \centering
    \includegraphics[width=\linewidth]{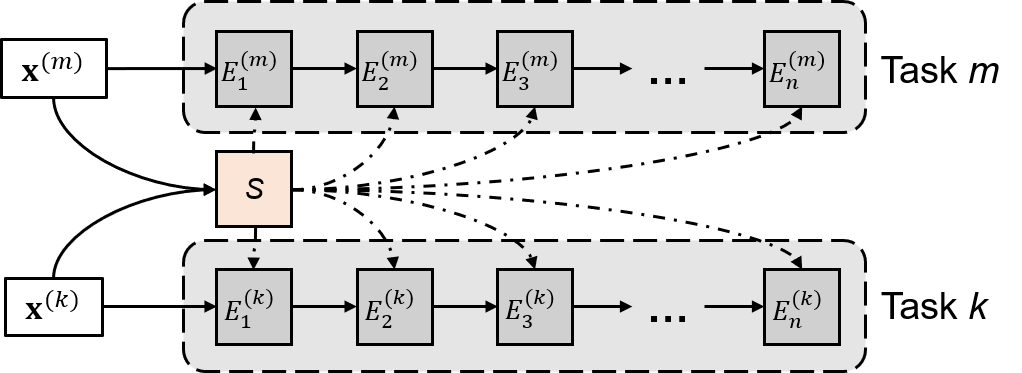}
    \caption{General global attention sharing architecture.}
    \label{fig:arc1}
\end{figure}

\begin{figure}[htbp]
    \centering
    \includegraphics[width=\linewidth]{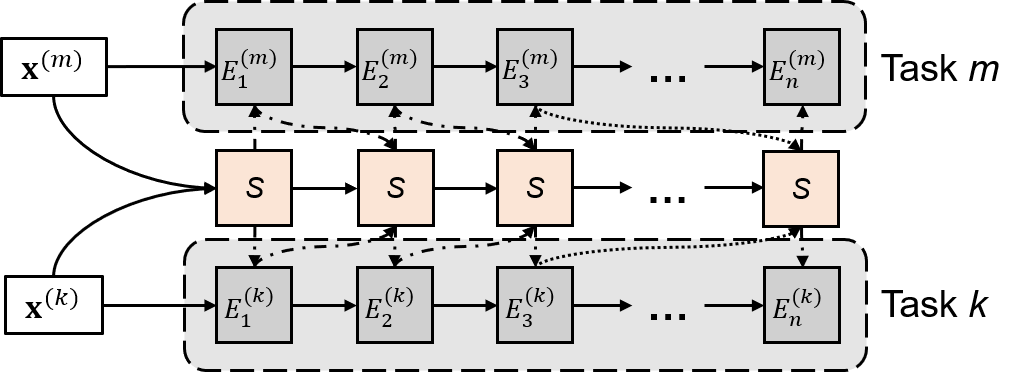}
    \caption{Hybrid attention sharing architecture.}
    \label{fig:arc2}
\end{figure}

By recurrently feeding outputs from task-specific encoders to the shared multi-head attention layer, this attention sharing architecture can enhance the capacity of memorizing while general global shared attention enables the information flowing from different tasks to interact sufficiently. Fig. \ref{fig:arc1} and Fig. \ref{fig:arc2}
clearly describe the two attention sharing architectures and illustrate the difference.  


\section{Experiments}
In this section, we investigate the empirical performances of our proposed architecture MTL-Trans on the following multi-task real-world dataset. 

\subsection{Dataset Description}

\textbf{TRA-MI} This trafﬁc dataset was published by \textsc{Telecom Italia}\footnote{ https://dandelion.eu/datamine/open-big-data/} and it contained network measurements in terms of total cellular trafﬁc volume obtained from Milan city in Italy, where the city was partitioned into $100 \times 100$ grids of equal size $235m \times 235m$. The measurements were logged over 10-minute intervals between 1 Nov 2013 and 1 Jan 2014. Interestingly, each divided area is regarded as an independent task while there are also some hidden connections between each area. As an example, region \textsc{A} and region \textsc{B} are geographically adjacent which means these two areas are somehow related, such as sharing similar geographic information or municipal resources. If our proposed model can learn the similarity between different tasks, there is no doubt it will enhance the generalization ability to forecasting other related tasks even without pre-training it. For computing efficiency, we geographically merge all the small grids into 10 regions as 10 different tasks. Each region contains 1000 samples and naturally be marked as Task$\#$1, Task$\#$2, etc.

\subsection{Benchmarks}
We extensively compare MTL-Trans to a wide range of models for time series forecasting. Hyperparameter optimization is conducted using random search over a pre-defined search space, using the same number of iterations across all benchmarks for the same given dataset. Specifically, for single-task learning, the methods in our comparative evaluation are as follows. 
\begin{itemize}
    \item \textbf{LSTM} \cite{Hochreiter1997} Recurrent neural network with two-layer hidden long-short memory units and dropout applied.
    \item \textbf{Seq2Seq-Attn} \cite{Sutskever2014, Bahdanau2015} Sequence to sequence network is a model consisting of two RNNs called the encoder and decoder. The encoder reads an input sequence and outputs a single vector, and the decoder reads that vector to produce an output sequence. Additionally, attention mechanism is applied. 
    \item \textbf{DeepAR} \cite{Salinas2019} Auto-regressive RNN time series model which consists of an LSTM that takes the previous time points and co-variates as input for next time step.
    \item \textbf{DSSM} \cite{Rangapuram2018} Deep state-space model is a probabilistic time series forecasting approach that combines state-space models with deep learning by parameterizing a per-time-series linear state-space model with a jointly-learned recurrent neural network.
\end{itemize}
For multi-task learning, we compare our proposed approaches with the RNN-based generic sharing schemes.
\begin{itemize}
    \item \textbf{SSP-MTL} \cite{Liu2016, Liu2016a} An LSTM-based multi-task sequence learning model with a shared-private sharing scheme by stacking hidden states from different tasks. 
\end{itemize}
For the single-task learning methods above, we trained each model on each task independently. All the models forecast one-time step forward with a consistent historical horizon. 

\subsection{Evaluation Metrics}
These methods are evaluated based on three commonly used metrics in time series forecasting, including:
\begin{itemize}
    \item Empirical Correlation Coefficient (CORR) 
    \begin{equation}
    \mathrm{CORR} = \frac{\sum_{t=1}^{n}\left(\hat{y}_{t}-\bar{\hat{y}}\right)\left(y_{t}-\bar{y}\right)}{\sqrt{\sum_{t=1}^{n}\left(\hat{y}_{t}-\bar{\hat{y}}\right)^{2}} \sqrt{\sum_{t=1}^{n}\left(y_{t}-\bar{y}\right)^{2}}}
    \end{equation}
    \item Root Mean Squared Error (RMSE)
    \begin{equation} 
    \mathrm{RMSE} = 
    \mathbb{E}\left[\frac{\sum_{t=1}^{n}\left(\hat{y}_{t}-y_{t}\right)^{2}}{n}\right]^{1/2}
    \end{equation}
    \item Symmetric mean absolute percentage error (sMAPE)
    \begin{equation}\mathrm{sMAPE}=\frac{100 \%}{n} \sum_{t=1}^{n} \frac{\left|\hat{y}_{t}-y_{t}\right|}{\left(\left|\hat{y}_{t}\right|+\left|y_{t}\right|\right) / 2}
    \end{equation}
\end{itemize}
where $y_{t}$ is the ground truth value and $\hat{y}_{t}$ is the forecast value.

\subsection{Training Procedure}
We partition all time series of all tasks into 3 parts in chronological order -- a training set (60\%) for learning, a validation set (20\%) for hyperparameter tuning, and a hold-out test set (20\%) for performance evaluation. All time series have been preprocessed by applying Min-Max normalization such that all the values range from -1 to 1. Hyperparameter optimization is conducted via random search, using 50 iterations. Additionally, we use AdamW optimizer \cite{Loshchilov2019} with learning rate decay strategy applied: the learning rate of each parameter group decayed by gamma $\gamma$ every pre-defined steps \footnote{All the experiments were done by using \texttt{Pytorch} library.}. Full search ranges for all hyperparameters are below, with optimal model parameters listed in Table. \ref{tab:configuration}. 
\begin{itemize}
    \item \textbf{Shared and each task-specific embedding dimension} -- 16, 32, 64, 128
    \item \textbf{Number of heads} -- 2, 4, 8
    \item \textbf{Number of encoder layers} -- 1, 2, 3, 4, 5, 6
    \item \textbf{Dimension of feed-forward layer} -- 128, 256, 512, 1024
    \item \textbf{Dropout rate} -- 0.1, 0.2, 0.3, 0.4, 0.5, 0.7, 0.9
    \item \textbf{Mini-batch size} -- 32, 64, 128, 256
    \item \textbf{Learning rate} -- 0.0003, 0.003, 0.03
    \item \textbf{Max. gradient norm} -- 0.01, 0.7, 1.0, 100.0
    \item \textbf{Learning rate decay rate} -- 0.80, 0.95, 0.99
    \item \textbf{Decay step size} -- 1.0, 5.0, 10.0
\end{itemize}

Following \cite{Collobert2008, Liu2016}, the training is achieved in a stochastic manner by looping over tasks:
\begin{enumerate}
    \item Randomly select a task $m$.
    \item Train a consecutive mini-batch $\mathbf{b}$ of samples from this task $m$.
    \item Update the parameters for this task by gradient backward with respect to this mini-batch $\mathbf{b}$.
    \item Go to Step 1.
\end{enumerate}

Across all training process, all task-specific models were trained on the same single NVIDIA Tesla P100 GPU, and can be deployed without the need for extensive computing resources. 

\subsection{Loss Function}
Both global-shared attention architecture and hybrid architecture are trained by minimizing the squared $L^2$ norm loss \cite{Lehmann1998}, summed across all outputs:
\begin{equation}
\ell(x, y)=L=\left\{l_{1}, \ldots, l_{N}\right\}^{\top}, \quad l_{n}=\left(x_{n}-y_{n}\right)^{2}
\end{equation}
where $N$ is the batch size. $x$ and $y$ are sequences of arbitrary shapes with a total of $n$ elements each.

\begin{table}[htbp]
\centering
\resizebox{.75\linewidth}{!}{%
\begin{tabular}{|l|l|}
\hline
                                   & \textbf{TRA-MI} \\ \hline
{\underline{\textbf{Dataset Details}}}     &        \\ 
Target Type                        & $\mathbb{R}$      \\ 
Number of Tasks                    & 10     \\ \hline
{\underline{\textbf{Network Parameters}}}  &        \\ 
Embedding Dimension                & 32     \\ 
Number of Heads                    & 4      \\ 
Number of Encoder Layers           & 2      \\ 
Dimension of FFN                   & 128    \\ 
Dropout Rate                       & 0.1    \\ \hline
{\underline{\textbf{Training Parameters}}} &        \\ 
Mini-batch Size                     & 64     \\ 
Learning Rate                      & 0.0003 \\ 
Max Gradient Norm                  & 0.7    \\ 
Learning rate decay rate           & 0.95   \\
Decay step size                    & 1.0    \\ \hline
\end{tabular}%
}
\caption{Information on dataset and optimal training configuration.}
\label{tab:configuration}
\end{table}

\subsection{Main Results}
\begin{figure}[htbp]
     \centering
     \includegraphics[width=\linewidth]{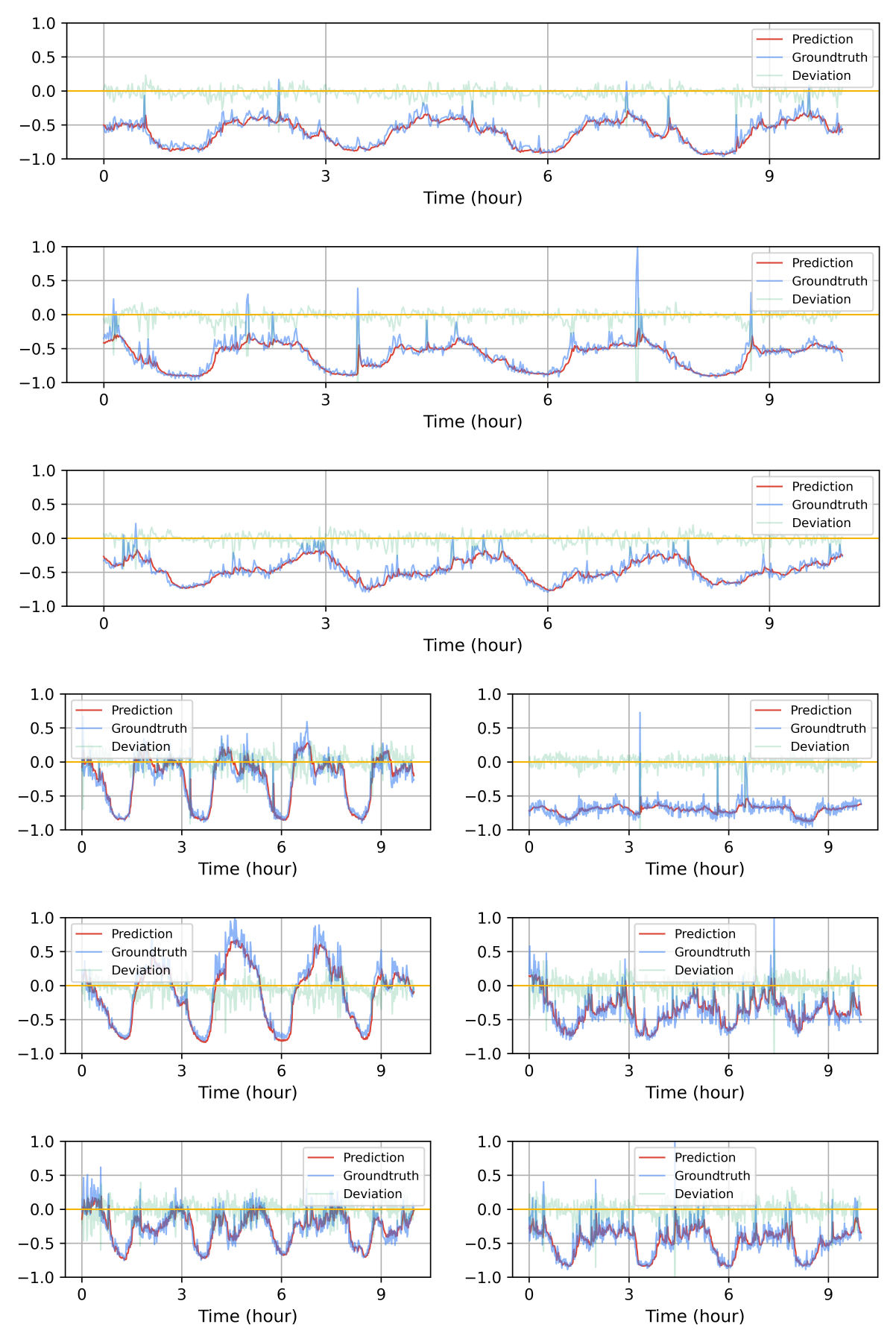}
     \caption{Predicted time series of nine randomly selected samples from different tasks with the same historical horizon as 15 hours and forecast window as 10 mins on test set.}
     \label{fig:prediction}
\end{figure}

\begin{table*}[htbp]
\centering
\resizebox{.9\textwidth}{!}{%
\begin{tabular}{@{}clcccccccc@{}}
\toprule
\multirow{2}{*}{Task} &
  \multicolumn{1}{c}{\multirow{2}{*}{Metrics}} &
  \multicolumn{4}{c}{Single-Task} &
  Multi-Task &
  \multicolumn{2}{c}{MTL-Trans (ours)} &
  \multirow{2}{*}{$\Delta$} \\ \cmidrule(lr){3-6} \cmidrule(lr){7-7} \cmidrule(lr){8-9} 
                      & \multicolumn{1}{c}{} & LSTM    & Seq2Seq-Attn & DeepAR  & DSSM    & SSP-MTL       & Global           & Local-Global     &         \\ \midrule
\multirow{3}{*}{$\#1$}  & CORR                 & 0.7108  & 0.8005       & 0.8536  & 0.8640  & {\ul 0.8885}  & 0.9045           & \textbf{0.9049}  & +1.85$\%$ \\
                      & RMSE                 & 0.1138  & 0.1050       & 0.0986  & 0.0979  & {\ul 0.0952}  & 0.0937           & \textbf{0.0934}  & +1.82$\%$ \\
                      & sMAPE                & 16.20$\%$ & 14.91$\%$      & 13.92$\%$ & 13.96$\%$ & {\ul 13.50$\%$} & 13.22$\%$          & \textbf{13.17$\%$} & +2.46$\%$ \\
\multirow{3}{*}{$\#2$}  & CORR                 & 0.6781  & 0.7673       & 0.8149  & 0.8248  & {\ul 0.8492}  & \textbf{0.8628}  & 0.8623           & +1.61$\%$ \\
                      & RMSE                 & 0.1279  & 0.1176       & 0.1102  & 0.1105  & {\ul 0.1068}  & \textbf{0.1044}  & 0.1046           & +2.25$\%$ \\
                      & sMAPE                & 14.66$\%$ & 13.45$\%$      & 12.53$\%$ & 12.51$\%$ & {\ul 12.17$\%$} & \textbf{11.96$\%$} & \textbf{11.96$\%$} & +1.74$\%$ \\
\multirow{3}{*}{$\#3$}  & CORR                 & 0.6967  & 0.7784       & 0.8392  & 0.8418  & {\ul 0.8679}  & \textbf{0.8836}  & 0.8835           & +1.81$\%$ \\
                      & RMSE                 & 0.1129  & 0.1041       & 0.0977  & 0.0966  & {\ul 0.0941}  & \textbf{0.0925}  & 0.0931           & +1.77$\%$ \\
                      & sMAPE                & 24.88$\%$ & 22.76$\%$      & 21.31$\%$ & 21.44$\%$ & {\ul 20.72$\%$} & \textbf{20.31$\%$} & 20.36$\%$          & +2.00$\%$ \\
\multirow{3}{*}{$\#4$}  & CORR                 & 0.7339  & 0.8301       & 0.8882  & 0.8901  & {\ul 0.9211}  & 0.9390           & \textbf{0.9392}  & +1.96$\%$ \\
                      & RMSE                 & 0.1488  & 0.1354       & 0.1276  & 0.1278  & {\ul 0.1236}  & 0.1211           & \textbf{0.1205}  & +2.50$\%$ \\
                      & sMAPE                & 83.24$\%$ & 76.10$\%$      & 71.75$\%$ & 71.68$\%$ & {\ul 69.42$\%$} & 68.16$\%$          & \textbf{68.15$\%$} & +1.83$\%$ \\
\multirow{3}{*}{$\#5$}  & CORR                 & 0.7585  & 0.8575       & 0.9183  & 0.9242  & {\ul 0.9489}  & 0.9638           & \textbf{0.9646}  & +1.65$\%$ \\
                      & RMSE                 & 0.1134  & 0.1045       & 0.0980  & 0.0974  & {\ul 0.0948}  & 0.0929           & \textbf{0.0921}  & +2.81$\%$ \\
                      & sMAPE                & 51.84$\%$ & 47.89$\%$      & 44.90$\%$ & 44.84$\%$ & {\ul 43.34$\%$} & 42.48$\%$          & \textbf{42.03$\%$} & +3.02$\%$ \\
\multirow{3}{*}{$\#$6}  & CORR                 & 0.6775  & 0.7618       & 0.8149  & 0.8199  & {\ul 0.8457}  & 0.8587           & \textbf{0.8593}  & +1.61$\%$ \\
                      & RMSE                 & 0.1635  & 0.1493       & 0.1405  & 0.1394  & {\ul 0.1356}  & 0.1331           & \textbf{0.1330}  & +1.98$\%$ \\
                      & sMAPE                & 70.18$\%$ & 64.76$\%$      & 60.65$\%$ & 60.15$\%$ & {\ul 58.63$\%$} & 57.31$\%$          & \textbf{56.89$\%$} & +2.98$\%$ \\
\multirow{3}{*}{$\#$7}  & CORR                 & 0.7572  & 0.8523       & 0.9131  & 0.9149  & {\ul 0.9444}  & \textbf{0.9593}  & 0.9580           & +1.58$\%$ \\
                      & RMSE                 & 0.0728  & 0.0667       & 0.0628  & 0.0626  & {\ul 0.0608}  & \textbf{0.0594}  & 0.0602           & +2.37$\%$ \\
                      & sMAPE                & 7.98$\%$  & 7.32$\%$       & 6.89$\%$  & 6.88$\%$  & {\ul 6.67$\%$}  & \textbf{6.52$\%$}  & 6.69$\%$           & +2.30$\%$ \\
\multirow{3}{*}{$\#$8}  & CORR                 & 0.5364  & 0.6001       & 0.6454  & 0.6494  & {\ul 0.6679}  & \textbf{0.6814}  & 0.6785           & +2.01$\%$ \\
                      & RMSE                 & 0.1051  & 0.0972       & 0.0910  & 0.0902  & {\ul 0.0880}  & \textbf{0.0865}  & 0.0868           & +1.77$\%$ \\
                      & sMAPE                & 9.75$\%$  & 8.99$\%$       & 8.40$\%$  & 8.41$\%$  & {\ul 8.15$\%$}  & \textbf{7.97$\%$}  & 7.99$\%$           & +2.28$\%$ \\
\multirow{3}{*}{$\#$9}  & CORR                 & 0.6369  & 0.7144       & 0.7659  & 0.7726  & {\ul 0.7969}  & 0.8107           & \textbf{0.8111}  & +1.78$\%$ \\
                      & RMSE                 & 0.1865  & 0.1712       & 0.1606  & 0.1602  & {\ul 0.1558}  & 0.1533           & \textbf{0.1532}  & +1.70$\%$ \\
                      & sMAPE                & 53.70$\%$ & 48.84$\%$      & 46.29$\%$ & 45.76$\%$ & {\ul 44.61$\%$} & \textbf{43.70$\%$} & 43.85$\%$          & +2.03$\%$ \\
\multirow{3}{*}{$\#$10} & CORR                 & 0.6457  & 0.7258       & 0.7706  & 0.7785  & {\ul 0.8035}  & \textbf{0.8231}  & 0.8227           & +2.44$\%$ \\
                      & RMSE                 & 0.1711  & 0.1571       & 0.1472  & 0.1468  & {\ul 0.1432}  & \textbf{0.1399}  & 0.1404           & +2.28$\%$ \\
                      & sMAPE                & 33.79$\%$ & 30.92$\%$      & 29.00$\%$ & 28.90$\%$ & {\ul 28.04$\%$} & 27.42$\%$          & \textbf{27.37$\%$} & +2.40$\%$ \\ \bottomrule
\end{tabular}%
}
\caption{Model performance of two proposed attention sharing schemes against state-of-the-art neural models on \textbf{TRA-MI} dataset. Best performance in boldface. $\Delta$ represents the improvements compared to SSP-MTL. Experiments on all tasks are with the same historical horizon as 15 hours and forecast window as 10 minutes. Our proposed MTL-Trans consistently outperform all benchmarks over the variety of tasks and metrics. }
\label{tab:main_results}
\end{table*}

\begin{figure}[htbp]
    \centering
    \includegraphics[width=\linewidth]{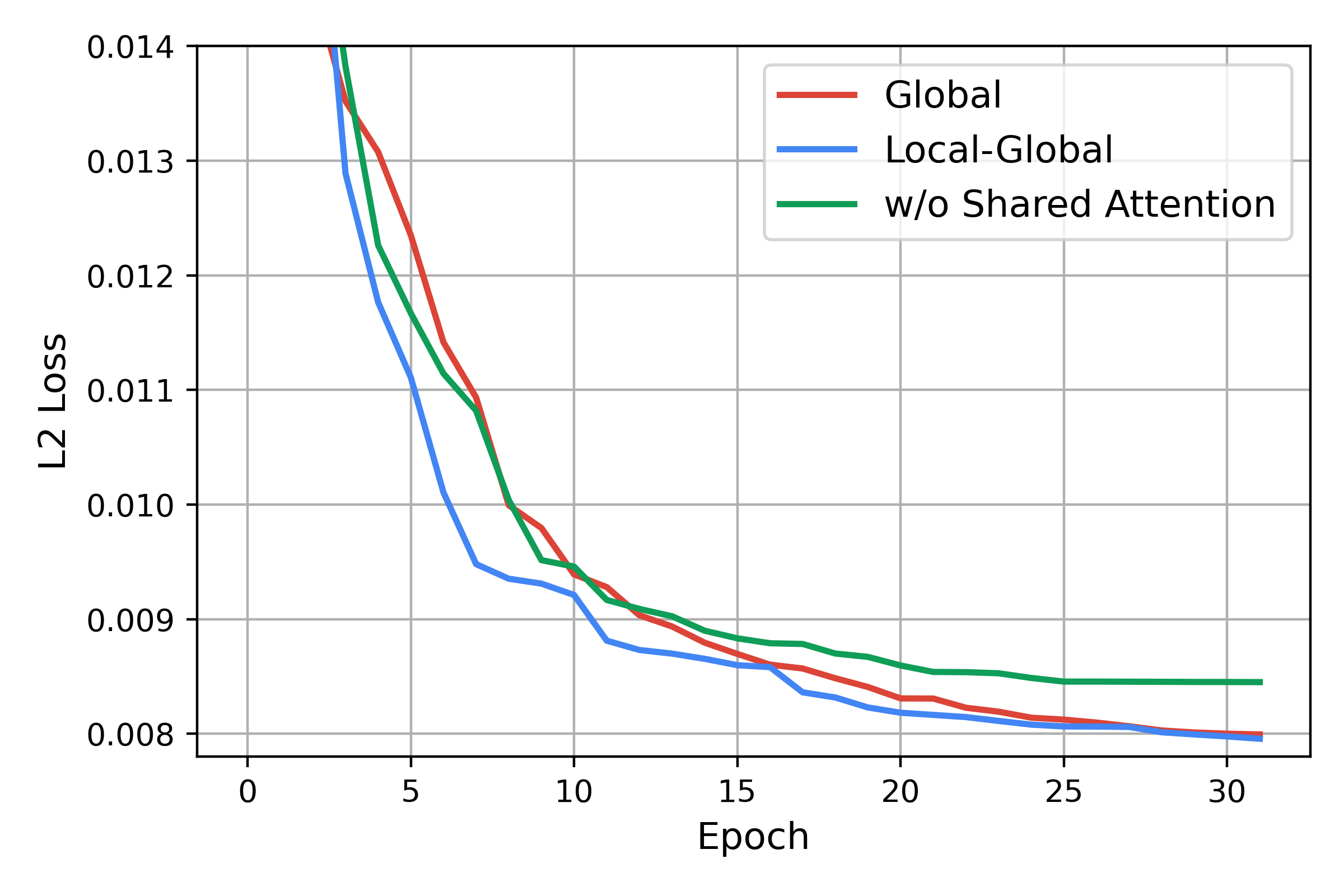}
    \caption{Loss decrement against training steps among three architectures.}
    \label{fig:ablation}
\end{figure}

We compare our proposed two architectures with a wide range of baselines for both single-task forecasting and multi-task forecasting. Table. \ref{tab:main_results} summarizes the forecasting performance of our proposed method through three evaluation metrics. Our proposed MTL-Trans architectures significantly outperform all benchmarks over the variety of metrics and tasks. Moreover, the multi-task frameworks (SSP-MTL \& ours) that jointly train the data outperform the single-task training framework as the model performance consistently tells. It demonstrates the shared information scheme across tasks can enhance modeling ability and capture both similarities and difference between tasks that finally benefits the model. With the help of the shared attention layer, the performances of all tasks by our proposed methods are significantly improved roughly around $2\%$ across all metrics compared to the LSTM-based architecture SSP-MTL. This consistent improvements demonstrate the long term dependency modeling capability of the self-attention mechanism. Moreover, the two different attention sharing schemes share the winning tickets since the global sharing scheme performs better on Tasks 2, 3, 7, 8, 10 while the other one performs better on the rest tasks. As we described in section \ref{section:scheme}, for tasks with highly similar patterns, a general global attention memory might be more suitable since consistent global attention helps capture the similarity and backward this information to each specific task more efficiently. For tasks with more inconsistent patterns, a local-global attention sharing scheme might be more appropriate because it can also record task-specific information besides globally shared information which can diversify each task-specific pipeline. 

\begin{figure}[htbp]
    \centering
    \includegraphics[width=\linewidth]{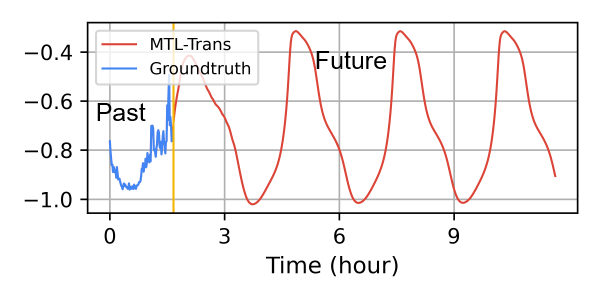}
    \caption{Simulation test on future prediction based on real data.}
    \label{fig:forecast}
\end{figure}

Fig. \ref{fig:prediction} shows some predicted time series by local-global sharing architecture. The predicted curve almost coincides with the groundtruth which further demonstrate the model's predictive capability. Fig. \ref{fig:forecast} further shows the good predictive ability of our proposed model as the simulated forecasting based on partial real data maintains the original pattern well. 

\subsection{Ablation Analysis}

One intuitive question is that what if we only train each task-specific transformer encoder separately instead of sharing public multi-head attention? If we tune the hyperparameter of each task-specific model (As an example, increase the number of heads, deepen the encoder layers, etc.) such that they own similar amount of model parameters to the shared-attention scheme, removing the performance gain induced by model complexity, will they perform better than our shared attention model? Fig. \ref{fig:ablation} tells us the answer by showing the loss decrement against training steps among three architectures -- global sharing scheme, hybrid local-global sharing scheme, and pure paralleled transformer encoders without sharing information.  
The fastest for loss descent is by local-global attention sharing scheme followed by the global attention sharing scheme and they eventually converged together. Compared to the shared attention architecture, the loss of pure encoders without sharing information drops more slowly, and the final result is not as good as the others which again demonstrates the effectiveness of sharing paradigm in multi-task learning. 

To further illustrate the effectiveness of MTL-Trans in modeling the multi-task time series data, we summarize the following reasons:
\begin{itemize}
\item First of all, there are similarities between all related tasks and one fundamental mission in multi-task learning is to find these similarities out and take further advantages of them to benefit in solving other unseen tasks. The shared attention captures the similarity between different tasks and feedback on all related tasks. This is the main reason why this shared attention architecture can outperform naive models. 
\item Self-attention mechanism is the second hero that helps to make this happen. As we have discussed in section IV, the essence of the self-attention mechanism is a soft-addressing process. Our shared multi-head attention plays an important role that helps to record this query-key pairwise addressing information that can benefit other unseen tasks under the hypothesis that similar tasks share similar self-addressing information. 
\end{itemize}

\section{Conclusion}
In this paper, we presented a shared attention-based architecture with two different sharing schemes for multi-task time series forecasting. By setting an external public multi-head attention function for capturing and storing self-attention information across different tasks, the proposed architectures significantly improved the state-of-the-art results in multi-task time series forecasting on this multi-resource cellular traffic dataset \textbf{TRA-MI}. With ablation analysis and empirical evidence, we show the efficiency of the proposed architecture and the essence of why it succeeds. For future work, we will investigate the following two aspects: (1) applying the proposed model to other sequence modeling tasks such as machine translation; (2) developing other attention sharing schemes to further enhance the predictive ability; (3) finding another way or architecture that computes the shared multi-head attention more efficiently, e.g. the time and memory complexity of computing a multi-head self-attention function would cost $\mathcal{O}(L^2)$ where $L$ is the length of input sequences. It could be hard to compute when the sequence length is very long or the computational power is limited.

\section*{Acknowledgment}
The authors would like to thank all the anonymous reviewers for their insightful comments. We thank Dr. Jian Pei for the discussion and constructive suggestion on the paper organization and experiments design. This work was partially supported by the US National Science Foundation under grant IIS-1741279.

\newpage
\bibliographystyle{IEEEtran}
\bibliography{reference}

\vspace{12pt}

\end{document}